\documentclass[a4paper,conference]{IEEEtran}
\usepackage{times}
\usepackage{epsfig}
\usepackage{graphicx}
\usepackage{amsmath}
\usepackage{amssymb}
\usepackage{mathtools} 

\usepackage[T1]{fontenc}
\usepackage{color}
\usepackage{xcolor}
\usepackage{nicefrac}
\usepackage{comment}
\usepackage{enumitem}
\usepackage{pgf} 
\usepackage{tabularx}
\usepackage{marvosym}
\usepackage{wrapfig}

\ifCLASSINFOpdf
\else
\fi
\newcommand\blfootnote[1]{%
  \begingroup
  \renewcommand\thefootnote{}\footnote{#1}%
  \addtocounter{footnote}{-1}%
  \endgroup
}

\begin{document}
%
\title{Gaining Scale Invariance in UAV Bird's Eye View Object Detection by Adaptive Resizing}

\author{\IEEEauthorblockN{Martin Messmer$^\dagger$}
\IEEEauthorblockA{Cognitive Systems Group\\
University of T\"ubingen\\
T\"ubingen, Germany\\
\Letter martin.messmer@uni-tuebingen.de}
\and
\IEEEauthorblockN{Benjamin Kiefer$^\dagger$}
\IEEEauthorblockA{Cognitive Systems Group\\
University of T\"ubingen\\
T\"ubingen, Germany\\
\Letter benjamin.kiefer@uni-tuebingen.de}
\and
\IEEEauthorblockN{Andreas Zell}
\IEEEauthorblockA{Cognitive Systems Group\\
University of T\"ubingen\\
T\"ubingen, Germany\\
\Letter andreas.zell@uni-tuebingen.de}}


%


\maketitle

\begin{abstract}
    This work introduces a new preprocessing step for object detection applicable to UAV bird's eye view imagery, which we call Adaptive Resizing. By design, it helps alleviate the challenges coming with the vast variances in objects' scales, naturally inherent to UAV data sets. Furthermore, it improves inference speed by two to three times on average. We test this extensively on UAVDT, VisDrone, and on a new data set we captured ourselves and achieve consistent improvements while being considerably faster. Moreover, we show how to apply this method to generic UAV object detection tasks. Additionally, we successfully test our approach on a height transfer task where we train on some interval of altitudes and test on a different one. Furthermore, we introduce a small, fast detector meant for deployment to an embedded GPU. Code will be made publicly available on our website.
\end{abstract}
\section{Introduction}
\label{sec:intro}
\blfootnote{$^\dagger$These authors contributed equally. This work has been supported by the German Ministry for Economic Affairs and Energy, Project Avalon, FKZ: 03SX481B}Deep learning-based research aimed at object detection has shown remarkable performance \cite{Lin2014, Redmon2016, Ren2015}. Despite these giant leaps in generic object detection, the particular case of UAV object detection, i.e. images taken from Unmanned Aerial Vehicles (UAVs), lags behind in terms of the best-performing models' accuracy on the most popular data sets \cite{wang2019learning}.
One of the main reasons for this discrepancy is the versatile application areas of UAVs with mounted cameras which lead to vast differences in the altitude above the ground of the UAV at the time of capture (capture-altitude). For example, in traffic surveillance applications, the altitudes can vary from 5 to 100 meters \cite{zhu2018visdrone}, while in search and rescue tasks, the span may be as large as 5 to 260 meters \cite{varga2022seadronessee}. This variance in altitudes results in a variance in objects' sizes. While humans are believed to have a scale-invariant perception and internal representation of objects \cite{han2020scale}, current object detectors do not. In fact, scale variation is a major cause for poor detection \cite{Singh2018}. While there is a corpus of works addressing this issue for generic object detection \cite{Singh2018,huang2019rotation,kokkinos2008scale,liu2019learning}, it remains a complicated problem to solve. 

On the other hand, in UAV bird's eye view object detection, objects' sizes mainly depend on the UAV's altitude. In turn, the altitude information is freely available via the UAV's onboard  barometer and GPS sensor. Current object detectors ignore this information entirely
. We argue that it is utterly helpful to include this valuable information as it tells us about the objects' sizes and how closely we have to look for objects. Analogous to humans' intrinsic understanding of their environment \cite{epstein2019scene}, we can incorporate that environmental information in the object detection pipeline to achieve a scale-invariant understanding of the scene.


\begin{figure}
	\begin{centering}
		\includegraphics[width=\linewidth]{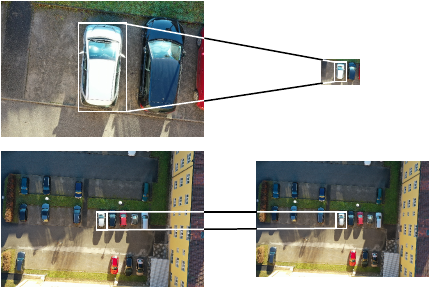}
		\caption{Example of the resizing process. On the left, we have two images from a possible UAV data set; the top one is captured at $10\, m$ flight altitude, the bottom one at $60\, m$. On the right, we again have both images resized according to their respective height. The bottom one stays roughly the same while the top one is resized by a large factor. Note how the bounding box of the silver car at the center of each respective picture is equal in size after resizing.}
		\label{fig:resize_example}
	\end{centering}
	\vspace{-6mm}
\end{figure}

Furthermore, ignoring the scale information of objects leads to models learning different representations of the very same objects if they are perceived at sufficiently different altitudes (and thus scales). In turn, this results in potential redundancy among the learned features. However, as onboard computation capabilities of UAVs are usually smaller than those of high-end consumer graphics cards, highly condensed models (with lower capacity) are needed.

Lastly, for higher altitudes, it is inevitable to provide large image resolutions to detect smaller objects \cite{varga2021tackling}. However, these large resolutions may be redundant in lower altitudes. Thus, an altitude-aware method benefits the inference time even further.







In this work, we tackle these problems by introducing a method we call Adaptive Resizer. At its core, this is a preprocessing technique designed to ensure that two arbitrary instances of the same class are of the same size throughout the entire data set. We do this by adaptively resizing each image depending on the altitude it has been captured in a principled way before passing it to an object detector.

This achieves two things: first, the object detector itself does not need to be scale-invariant. Second, the inference is much quicker because images taken at low altitudes are downscaled by a significant factor because they feature the largest objects. 

Our approach works for the special case of bird's eye view (BEV) images, i.e. images facing directly downwards, which form the most challenging subset \cite{Wu2019}. However, we also show the usefulness in general UAV object detection. To summarize, our key contributions are as follows: 

\begin{itemize}
	\item We propose a novel height-adaptive image preprocessing method, which improves UAV bird's eye view object detection performances in both accuracy and inference speed and is applicable to all state-of-the-art object detectors.
	\item We construct a fast object detector for embedded applications that builds upon this method.

\end{itemize}


\section{Related work}
\label{sec:related}

Object detectors can broadly be divided into two categories; one-stage and two-stage detectors. Two-stage detectors \cite{Ren2015, Girshick2015} are generally more accurate and therefore occupy the first places on established leader boards \cite{Du2019}. However, their inference speed is generally a lot lower than that of one-stage detectors \cite{Redmon2016, Zhou2019, Tian2019, Lin2017}, which makes the latter more suitable for onboard object detection scenarios.
Most recently, there are also transformer-based object detectors performing very well in generic object detection \cite{liu2021swin, zhu2020deformable, carion2020end}. They have, however, not proven to be useful for UAV or BEV object detection so far. \\
The closest method to ours is \cite{Kim2020}. There, images are also resized in accordance with the height. However, the authors resize every image to the same resolution (an average over the data set) while we calculate an individual size for each image. Furthermore, they merely test their method on class agnostic detection tasks. \\
While the authors in \cite{Singh2018} analyze the problem of scale invariance in CNN's in great depth, their solution employs an image pyramid, which is not viable for real-time detection. 
Another approach is presented in \cite{Yang2019}, where the authors try to detect clusters of potential targets and then predict the scale offset before regressing the objects in each cluster more accurately. A drawback is the need for ground truth labels of clusters. Furthermore, the sequential use of multiple different networks is computationally expensive, while our approach estimates scales for the whole image deterministically. \\
Most papers tackling real-time object detection in general \cite{Redmon2016, redmon2018yolov3} or on mobile platforms \cite{Ringwald2019} design a whole network architecture. Meanwhile, this paper introduces a method applicable to most modern object detectors, improving their speed and detection performance. \\
The authors of \cite{Wu2019} propose to apply adversarial learning techniques to the meta-data of UAV imagery. While they achieve good results, they only use the meta-data during training and not during validation. Also using it at test time can improve performance even further, as we show. \\
One recent work exhaustively examines how feature pyramid networks work and how object detectors (don't) benefit from them \cite{chen2021you}. However, compared to their approach we can choose a rather simple method to cut the feature pyramid network and therefore save on computational cost. That is, because the approach in \cite{chen2021you} aims at generic object detection while we go for the special case of BEV object detection.
\vspace{-1mm}

\section{Method}
\label{sec:method}

\vspace{-1mm}
The \textit{Adaptive Resizer} is a preprocessing strategy designed to address bird's eye view (BEV) object detection, i.e. object detection from UAVs where the angle of view is pointing downwards in a right angle. The Adaptive Resizer rescales every image in a principled manner to diminish the scale variance problem in BEV object detection.

One problem in BEV object detection 
is that object instances of the same class appear in vastly different sizes; see, for example, the left two images in Figure \ref{fig:resize_example}. This scale variance is primarily attributed to the altitude an image is captured at (\textit{capture-altitude}). 
A vanilla object detector is not aware of the fact, that it observes instances of the same class (or even the same object) but at different scales \cite{Lin2016}. Therefore, it learns different representations for different scales of the same object. That means some of the capacity of the detector is wasted on learning these different representations. One could either make use of this capacity in a different way or use a smaller object detector to increase inference speed.
Furthermore, an object detector that can make use of differently scaled training samples of the same objects makes more efficient use of the training samples. \\
However, the advantage of UAV object detection is the availability of freely available meta-data generated by the UAV during flight. That includes data like the camera's angle, capture-altitude, or time-stamp. The necessary meta-data for the Adaptive Resizer is the capture-altitude.
Unique to BEV object detection is that all instances of the same class are roughly equal in size on any single image, because all objects are about the same distance from the camera.


Building on that, the Adaptive Resizer achieves its goal (making every object of one class of same size over full data set) by resizing each image according to its height.
For this, the relevant determinant is the

\subsubsection*{Ground Sample Distance (GSD)}
\begin{wrapfigure}{r}{0.25\textwidth}
   \centering
   \vspace{-9mm}
   \begin{tabular}{@{}c@{\hspace{.5cm}}c@{}}
       \includegraphics[width=.2\textwidth, angle=0]{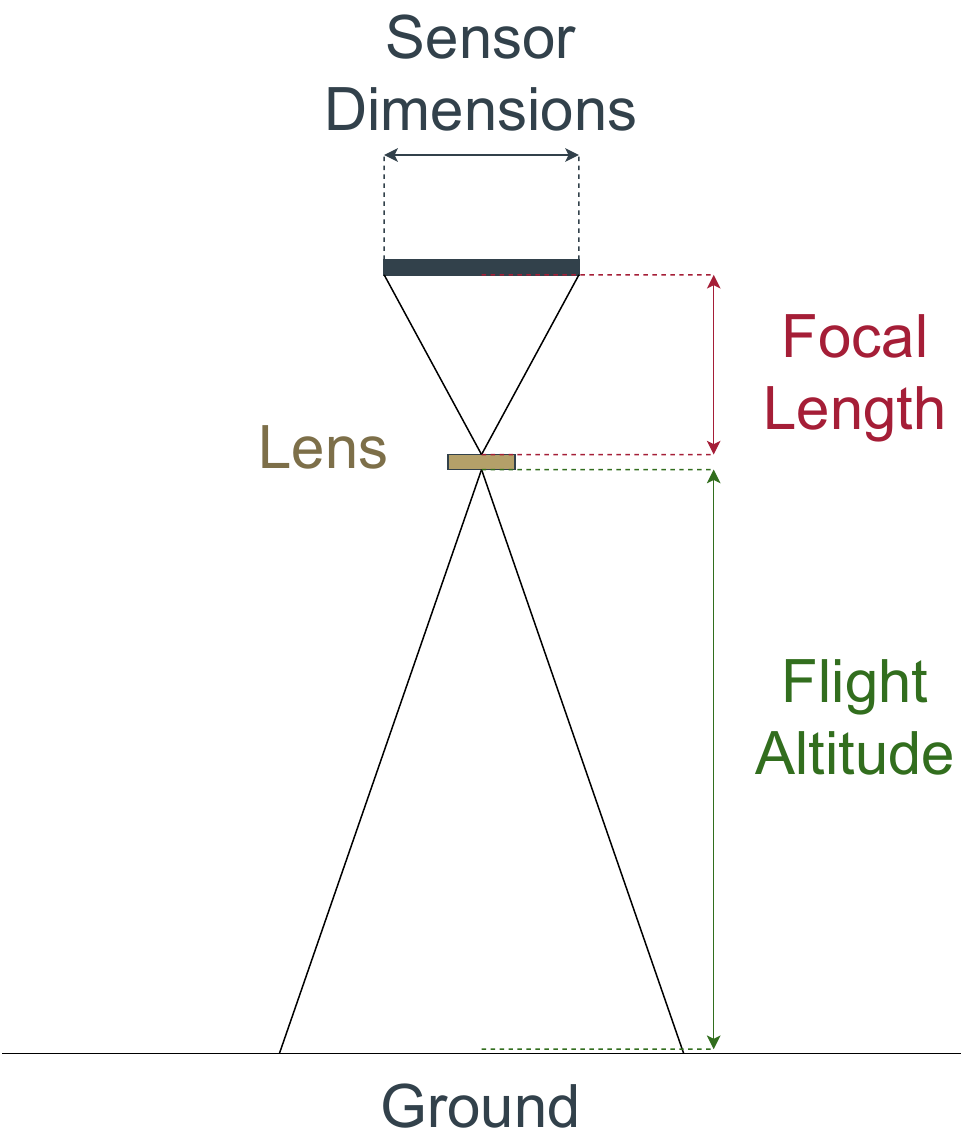} \\
   \end{tabular}
   \vspace{-3mm}
 \caption{Pictogram of a camera setup mounted on a UAV.}
 \label{fig:GSD-pictogram}
 \vspace{-3mm}
\end{wrapfigure}

To define the GSD of an image, let $p$ be its centre pixel. The definition of the GSD is the side length of the area on the ground that $p$ depicts.
For the calculation of the GSD we assume a fixed camera setup on the UAV. We can readily deduce the following formula from fundamental properties of the camera geometry (see Figure \ref{fig:GSD-pictogram}). \begin{align}
	\label{eq:GSD_eq}
	\operatorname{GSD} = \frac{S}{L} \cdot \frac{A}{I} . 
\end{align}
$S$ refers to the optical sensor's side length, while $A$ denotes the capture-altitude. $L$ refers to the camera's focal length, and $I$ denotes the captured image's side length. With a fixed camera setup, the only varying factors in Equation (\ref{eq:GSD_eq}) are the distance above ground $(A)$ and the image size $(I)$. Therefore, if we ensure that the ratio $\nicefrac{A}{I} \eqqcolon C$ is constant over the data set, the GSD is also constant across the entire data set. 
Ensuring that the GSD is constant over the data set is just a reformulation of the Adaptive Resizer's objective to alleviate the scale variance problem within each class. 

To implement adaptive resizing and make use of Equation (\ref{eq:GSD_eq}), we need to fix a reference class from the data set to determine the desired GSD, e.g. 'car'. Also, we fix a reference area, which is the goal size for all objects of the reference class after resizing.
Then, there are two ways; ideally, we know how large the standard representative of this reference class is. For example, if we fixed 'car' and know that the average car in the data set is $4 \, m \times 2 \, m$ while our reference area is \hbox{$32 \text{ px} \times 32$ px}, we get the desired GSD in two easy steps: \\
First, we compute the reference area with the same aspect ratio as the average car. Here, this is roughly $45.25 \text{ px} \times 22.63 \text{ px}$. Then the desired GSD is $\frac{4}{45.25} \nicefrac{m}{\text{px}}$. If we plug that into Equation (\ref{eq:GSD_eq}), we get the image size to resize to by solving for $I$. \\
If we do not know the size of the average car in our data set, we can still apply the Adaptive Resizer. In this case, we compute the average area of the bounding boxes of our reference class for a given image from the data set. Then we resize the image for this average to match our reference bounding box size. So if $\Tilde{I}$ is the size of the image, $M$ is the mean over the bounding box areas, and $R$ is the reference area, the image size to resize to is computed by
\begin{align}
\vspace{-5mm}
    \label{eq:Image_size_equation}
    I = \frac{R}{M} \cdot \Tilde{I} .
\vspace{-5mm}
\end{align}
The second method, taking the image-wise means of the bounding boxes, works consistently. However, the first method is more desirable as it filters annotation mistakes. Also, the second method does not work for images without instances of the reference class. \\
For an illustration of the whole process, see Figure \ref{fig:resize_example}. {This method works together with any modern deep learning-based object detector since our approach is a preprocessing step.}

Lastly, we note that we disregard effects of lens distortion and perspective projection as these are minor compared to the general relation of altitude to object size.

\subsection{{Building a Detector for Embedded Deployment}}
\label{subsec:fast_detector}

In this section we will leverage the new features the Adaptive Resizer brings to an object detector to build a fast detector for BEV imagery meant for embedded use.
We start with an EfficientDet--$D0$ \cite{Tan2019Effdet} in order to have a fast state-of-the-art detector and then omit the parts that we argue are not necessary in combination with adaptive resizing. EfficientDet is a family of models which are building onto EfficientNet-backbones \cite{Tan2019} and are therefore scalable in parameters, ranging up to EfficientDet--$D7$. Here, a higher number means that the model is larger and more accurate, while a lower number means that it is faster. We choose this detector because it is the smallest representative of its family, which in turn is the current AP50-state-of-the-art on COCO \cite{paperswithcode}.

EfficientDet--$D0$ employs a Feature Pyramid Network (FPN) \cite{Lin2016}, as is standard for modern object detectors (which are not transformer-based \cite{zhu2020deformable, carion2020end}). The FPN aims at making the detector perform well on multiple different levels of scale, because Convolutional Neural Networks (CNNs) are not inherently scale-invariant \cite{Singh2018}. 
An FPN extracts feature representations from the backbone network at different levels of depth, see Figure \ref{fig:EffDet}. Deeper ones are responsible for detecting larger objects because of their bigger field view (FOV), while earlier ones are being used to detect smaller objects. This is usually realized by distributing a vast number of prior boxes, called anchor-boxes, each corresponding to one feature map from the FPN. An anchor-box corresponding to a feature level of the FPN means, that the head from this feature level is used to classify and regress this anchor-box.
In the case of EfficientDet, the FPN employs five different feature levels. These levels are responsible for detecting objects at exponentially increasing sizes; EfficientDet uses 
$( 32, 64, 128, 256, 512)$. Therefore EfficientDet's anchor-boxes are of these sizes.

While this is an appropriate choice for data sets featuring everyday objects like COCO \cite{Lin2014} or Pascal VOC \cite{everingham2015pascal}, in BEV object detection, four out of these five feature levels are almost unused for each given image, see Table \ref{tab:featuremaps_adaptive}. This is due to the object sizes the respective feature maps are looking for and because in one given image from the BEV portion of a UAV data set all objects of a given class are (roughly) equal in size.

However, the network itself does not need to be scale-invariant, if all the objects in the data set are of the same size. For BEV images, all instances of any given class on one single image are a priori roughly equal in size, because all of them are about the same distance from the camera. Consequently, the only remaining problem is the objects' difference in scale between different images, precisely what the Adaptive Resizer aims at.

Consequently, we eliminate the feature pyramid network (FPN) from our model and only use the earliest feature map of those extracted from the backbone network. For an EfficientDet--$D0$ this reduces the number of parameters from around $4$m to roughly $0.5$m. This also leads to a large boost in inference speed, see section \ref{sec:experiments}.

\begin{figure}
	\begin{center}
		\scalebox{0.5}{\input{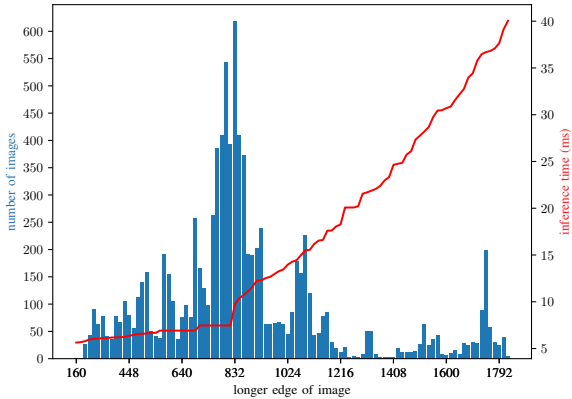}}
	\end{center}
	\vspace{-5mm}
	\caption{Distribution of image sizes after applying Adaptive Resizer on the UAVDT data set and the resulting inference time. The $x$-axis denotes the longer respective edge of the image, aspect ratios are kept during this process. The $y$-axis denotes the quantity in blue and the inference time in red.}
	\label{fig:size_distributionUAVDT}
	\vspace{-3mm}
\end{figure}

\section{Experiments}
\label{sec:experiments}

We employ Faster R-CNN \cite{Ren2015}, CenterNet \cite{zhou2019objects}, and EfficientDet--$D0$ \cite{Tan2019Effdet} to test our approach. We chose these three to have experiments with representatives of multiple major classes of object detectors. The first is a well known two-stage detector which is highly adjustable, for example with different ResNet-\cite{he2016deep} or ResNeXt\cite{xie2017aggregated}-backbones. The latter two are well-known one-stage detectors. EfficientDet is an anchor-based object detector while CenterNet is an anchor-free object detector \cite{Zhang2020}.

In the following, we will always report AP$^{50}$ values, as is usual for UAV data sets, except where explicitly stated otherwise. 

\subsection{Results on bird's eye view Portions}
\label{subsec:birdview}

We conduct our experiments on two well-known UAV data sets, VisDrone \cite{Zhu2018} and UAVDT \cite{Adaimi2020}, and on a new data set we recorded ourselves, which will be made publicly available. It is called \textbf{P}eople \textbf{O}n \textbf{G}rass (POG). The two former consist of around $7$k and $40$k images, respectively, and were both captured in major Asian cities. The latter contains roughly $2.8$k images, mostly showing people on a grass background. We captured POG to test our approach on because it features accurate height information per image, a very rare quality among UAV data sets.
As mentioned earlier, we conduct our experiments on each data set's BEV portion. These subsets contain roughly $1.4$k, $9.4$k, and $1$k images, respectively. Following the original authors of UAVDT, we combine all classes of their bounding box annotations into the single class `car` for our experiments due to heavy class imbalances. Because the existing altitude annotations are too coarse for our purposes, we generate finer height data artificially for UAVDT and VisDrone.
We do so using the second method from Section \ref{sec:method}. More precisely, we generate the image sizes by Equation (\ref{eq:Image_size_equation}). For POG, we extracted the log files from the UAV. Therefore, the data set contains meta annotations for each image, particularly altitude information, that is accurate to within one meter \cite{webinar2020gps}. The data set contains images in between $10\, m$ and $110 \, m$. 

For the experiments on one-stage detectors, we employ EfficientDet--$D0$ and CenterNet as described in their respective original papers \cite{Tan2019Effdet, zhou2019objects}. In the case of EfficientDet we fine-tuned hyper parameters like image size and anchor parameters (scales and ratios) to each data set. For CenterNet we did the same, except that it is anchor-free and therefore does not have anchor parameters.
To test our approach, we also do experiments with both networks employing the Adaptive Resizer.
We report the results of these experiments in Table \ref{tab:resultsOneStage} and \ref{tab:resultsCenterNet}. For both models we observe that employing Adaptive Resizer improves inference speed by a factor of two to three, see also Section \ref{subsec:benchmarks}.
In the case of EfficientDet (Table \ref{tab:resultsOneStage}) we can see that employing adaptive resizing achieves roughly an improvement of $3$ points AP$^{50}$ for VisDrone and POG. For UAVDT it even boosts performance by around $25$ points AP$^{50}$. See below for a discussion of this large gap in performance increase. 
For CenterNet (Table \ref{tab:resultsCenterNet}) the models employing Adaptive Resizer consistently outperform their baseline counterparts. On UAVDT in the most extreme case even by $28$ points AP$^{50}$. On VisDrone, however, the Adaptive Resizer only performs competitively with the baseline. \\
\indent We also include results for the Adaptive Resizer on two-stage detectors. While these are not relevant for onboard processing, they are still the most capable object detection models. For UAVDT we employ the baseline from \cite{Wu2019} to compare with their approach, as they are also using meta-information like capture-altitude. It is a Faster R-CNN network with Resnet-101-FPN backbone \cite{he2016deep}. For VisDrone, we reimplemented DE--FPN, which is the best-performing single model of the VisDrone Detection Challenge \cite{zhu2018visdrone}. We achieved $49.0$ AP$^{50}$ on the full validation set compared to their $49.1$ AP$^{50}$ on the full test set.
To compare it to our model, we train and test it on the BEV portion, then employ this as the baseline (in both cases). From the results in Table \ref{tab:resultsTwoStage} we observe that employing the Adaptive Resizer improves detection results for both data sets. While we improve by $5$ AP$^{50}$ points on VisDrone, we even achieve an improvement of over $13$ AP$^{70}$ on UAVDT compared to our baseline. We use the AP$^{70}$ metric to compare our approach to \cite{Wu2019} and observe that our model outperforms theirs by around $4$ points.

Summarizing all experiments, we observe that the Adaptive Resizer increases detection performance in general. However, the gain in performance is most prominent on UAVDT. We argue that this is due to the bad distribution of capture-altitudes in this data set. We observed that capture-altitudes are on average a lot lower in the training set of UAVDT than in its test set (which is not the case for VisDrone and POG). These are conditions the Adaptive Resizer can cope with very well, while generic object detectors suffer greatly, see Section \ref{subsec:domainadaptation}. Additionally, employing adaptive resizing speeds up inference by a factor of two to three. 

\begin{table}[h]
	\centering
	\begin{tabular}{c || c | c | c | c}
		& VisDrone & UAVDT & POG & FPS \\\hline\hline
		$D0$ @ 2048 & $13.1$ & $34.1$ & $80.3$ & $12$ \\\hline
		$D0$ @ 1792 & $17.7$ & $30.0$ & $74.3$ & $15$ \\\hline
		$D0\, + \,$Adaptive & $\mathbf{20.6}$ & $\mathbf{58.8}$ & $\mathbf{83.0}$ & $\mathbf{32}$
	\end{tabular}
	\caption{AP$^{50}$ results on the bev portions of the data sets. EfficientDet--$D0$@$x$ is a baseline model trained and evaluated such that the longer edge of each image is equal to $x$. 
	All FPS values are benchmarked on UAVDT and an RTX $2080$ ti GPU.}
	\label{tab:resultsOneStage}
\end{table}

\begin{table}[h]
	\centering
	\begin{tabular}{c || c | c}
		& UAVDT & VisDrone \\\hline\hline
		Faster R--CNN & $23.0$ & $41.0$ \\\hline
		NDFT \cite{Wu2019} & $32.9$ & -- \\\hline
		Adaptive & $\mathbf{36.8}$ & $\mathbf{46.0}$
	\end{tabular}
	\caption{AP$^{70}$ results on the bev portions of UAVDT and AP$^{50}$ on VisDrone. We use the AP$^{70}$ metric to compare our approach with \cite{Wu2019}.}
	\label{tab:resultsTwoStage}
\end{table}


\begin{table}[h]
	\centering
	\begin{tabular}{c || c | c | c}
		& UAVDT & VisDrone & FPS \\\hline\hline
		CN--RN18 Baseline & $33.7$ & $\mathbf{23.7}$ & $20$ \\\hline
		CN--RN18 Adaptive & $\mathbf{56.8}$ & $22.1$ & $\mathbf{55}$ \\\hline\hline
		
		CN--RN50 Baseline & $35.4$ & $\mathbf{28.5}$ & $8$ \\\hline
		CN--RN50 Adaptive & $\mathbf{63.4}$ & $26.3$ & $\mathbf{23}$ \\\hline\hline
		
		CN--RN101 Baseline & $37.6$ & $26.3$ & $5$ \\\hline
		CN--RN101 Adaptive & $\mathbf{60.8}$ & $\mathbf{26.5}$ & $\mathbf{6}$ \\
	\end{tabular}
	\caption{AP$^{50}$ results and frames per second (fps) of different CenterNet-models\cite{Zhou2019}. They differ in their respective backbone, for example 'CN--RN101' is a CenterNet with a ResNet101\cite{he2016deep} backbone.}
	\label{tab:resultsCenterNet}
\vspace*{-07mm}
\end{table}

\subsection{Effects of Cutting FPN}
\label{subsec:FPNvsnoFPN}

In this section we compare the detector from Section \ref{subsec:fast_detector}, meant for fast inference and deployment to an embedded GPU, to a full-fledged EfficientDet--$D0$ model with Adaptive Resizer. They differ in the fact that the model from Section \ref{subsec:fast_detector} has no feature pyramid network and therefore only uses one feature map. Table \ref{tab:featuremaps_adaptive} provides empirical evidence for that measure.
There, we can see the mean percentage of objects per image, that are detected by each feature map. 
Being detected by a certain feature map means, that the anchor which is selected to classify and regress the object in question (see description in Section \ref{subsec:fast_detector} and Figure \ref{fig:EffDet}) is corresponding to this specific feature map.
We discriminate between the detection percentage by feature map before and after applying non-maximum suppression (NMS). The values before NMS give an undistorted view of which feature maps would in principle be able to detect an object. The numbers after NMS, however, are more relevant to the application in practice, because only here does the detector filter predictions with poor scores; these are usually the ones which also regress the object worse than others.
In Table \ref{tab:featuremaps_adaptive} we can see that, after applying non-maximum suppression, on average less than two percent of all object per image are not detected by the first feature map. \\

\begin{table}[h]
	\centering
	\begin{tabular}{c || c | c | c | c}
		& 1 & 2 & 3 & 4 \& 5 \\\hline\hline
		pre NMS & $92.03$ \% & $7.95$ \% & $0.03$ \% & $0.00$ \%\\\hline
		post NMS & $98.01$ \% & $1.97$ \% & $0.02$ \% & $0.00$ \%
	\end{tabular}
	\caption{Average number of objects that are detected by each feature map before and after applying non-maximum suppresion (NMS). The average is taken over UAVDT. The investigated model is an EfficientDet--$D0$ with FPN and Adaptive Resizer.}
	\label{tab:featuremaps_adaptive}
\end{table}

\begin{table}[h]
	\centering
	\begin{tabular}{c || c | c | c | c}
		& UAVDT & VisDrone & POG & FPS \\\hline\hline
		$D0$--noFPN & $49.3$ & $\mathbf{23.6}$ & $79.2$ & $\mathbf{56}$ \\\hline
		$D0$--FPN & $\mathbf{58.8}$ & $20.6$ & $\mathbf{83.0}$ & $32$
	\end{tabular}
	\caption{AP$^{50}$ results on UAVDT, VisDrone, and POG. The compared models are EfficientDet--$D0$ with Adaptive Resizer. $D0$--noFPN is a model without FPN like described in Section \ref{subsec:fast_detector}, $D0$--FPN is the standard model with Adaptive Resizer, including fpn.}
	\label{tab:resultsFPN}
\end{table}

\subsection{Results on the complete UAVDT data set}
\label{subsec:resultsUAVDT}
To also introduce a model that works on a full UAV data set, we use a multi-domain approach in the style of \cite{kiefer2021leveraging}. More explicitly, we use the meta-data supplied by the UAV to distinguish between bird's eye view images and non-bird's eye view. During inference, we use the Adaptive Resizer model on the BEV images and a baseline model on all other images. Both are loaded before inference and available in GPU memory, so there is little overhead added and no drop in inference time for each of the models.
To achieve the results reported in Table \ref{tab:expertAdaptive} on UAVDT we use the models from Table \ref{tab:resultsTwoStage} for the two-stage detector experiments. For the experiments with EfficientDet--$D0$ we use the model without FPN from Section \ref{subsec:fast_detector} and the baseline from Table \ref{tab:resultsOneStage}.

\makeatletter
\def\@captype{figure}
\makeatother
  \includegraphics[width=0.45\textwidth, angle=0]{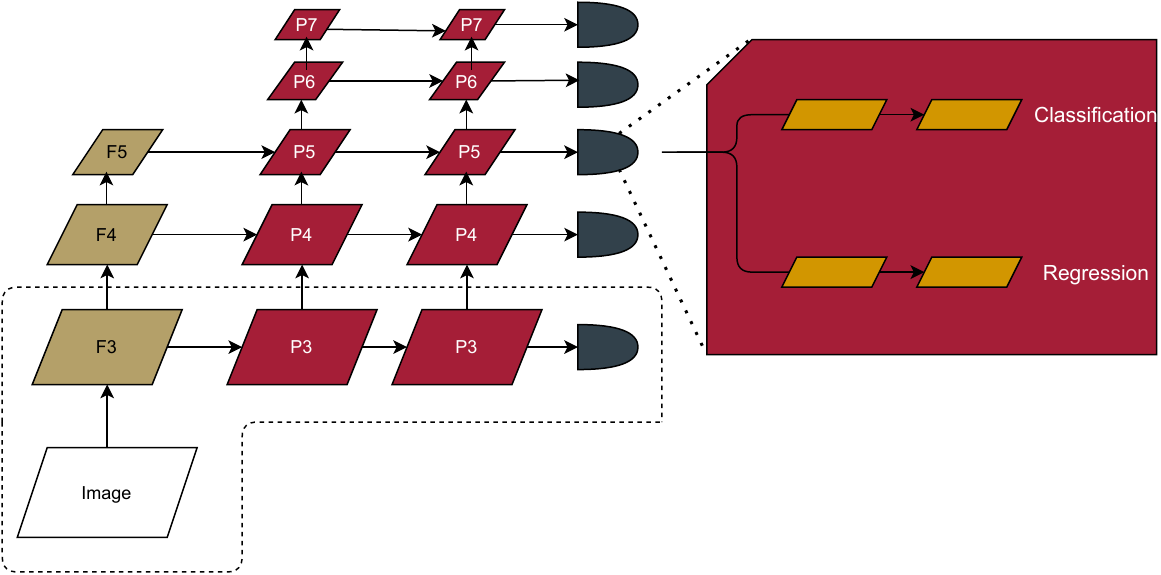}
  \caption{Schematic drawing of an EfficientDet--$D0$. Starting from the image, the backbone-network extracts feature maps $F3,F4,F5$ (gold). Then these are input to the feature pyramid network $P3-P7$ (red) and afterwards handed to the heads (anthracite). These perform classification and regression. The object detector without FPN from Section \ref{subsec:fast_detector} is encircled with the dashed line.}
  \label{fig:EffDet}

\vspace{2mm}
We observe that both models improve by circa $3$ AP points. Note that we are en par with \cite{Perreault2020}, also achieving $52.8$ AP$^{70}$. They give, to the best of our knowledge, the state-of-the-art detector on UAVDT. However, they employ a vastly more complicated method which needs short video sequences to perform well.

\begin{table}[h]
\centering
	\begin{tabular}{c || c | c}
		& Faster R--CNN & $D0$ \\\hline\hline
		Baseline & $49.4$ & $34.6$ \\\hline
		SpotNet \cite{Perreault2020} & $\mathbf{52.8}$ & -- \\\hline
		Adaptive & $\mathbf{52.8}$ & $\mathbf{37.7}$
	\end{tabular}
	\caption{Results on the full UAVDT data set. We use the AP$^{70}$ metric to compare our approach with the reported numbers in \cite{Perreault2020}.
	}
	\label{tab:expertAdaptive}	
\end{table}

\subsection{Time benchmarks}
\label{subsec:benchmarks}
Tables \ref{tab:resultsOneStage} and \ref{tab:resultsCenterNet} show that the Adaptive Resizer makes a model two to three times faster than its respective baseline. The reported number is the average of the inference times over the UAVDT BEV data set. We take the mean because the inference time for an Adaptive Resizer model is not constant; like for every object detector the inference time is dependant on the image size, which in turn is dependant on the capture-altitude.
We chose UAVDT to average over because it is the largest of the data sets we tested on. Therefore, it is the least prone to statistical outliers during the benchmark test.
Table \ref{tab:resultsOneStage} and \ref{tab:resultsFPN} show, where the speed improvement of the EfficientDet--$D0$+Adaptive Resizer without FPN comes from. Cutting the FPN from the model brings an improvement of $24$ FPS, which is larger than expected. We argue that this is due to the convolutional layers in the FPN, especially in later layers, having higher channel-dimensions than earlier layers \cite{Tan2019Effdet}. Because convolutional networks are essentially fully-connected in the channel-dimension, cutting these brings the largest speed improvement.

Figure \ref{fig:size_distributionUAVDT} explains the speed improvement when using Adaptive Resizer without any other alterations. All images captured at low altitudes are resized to comparably small image sizes, speeding the network up a lot, while the baseline runs at constant speed.
One could argue that the speed comparison is not fair because the baseline is employing a larger image size. However, this is necessary; otherwise, the baseline's AP deteriorates (as we saw in experiments) because of the small objects in UAV data sets \cite{Singh2018, varga2021tackling, Unel2019}.

We also benchmarked our EfficientDet--$D0$ with Adaptive Resizer and without FPN on a Jetson AGX Xavier development board optimized with \texttt{TensorRT} and half-precision FP$16$. There, our model achieved roughly $16$ FPS averaged over UAVDT. Meanwhile, the baseline achieved $7$ and $5$ FPS when resizing the image's longer respective side to $1792$ and $2048$ px, respectively, as in Table \ref{tab:resultsOneStage}. Therefore, on embedded hardware, Adaptive Resizer improves inference speed by a factor of two to three.

\subsection{Height Transfer}
\label{subsec:domainadaptation}

Without Adaptive Resizing, a model learns different representations for object instances with varying scales, as discussed in Section \ref{sec:method}.
Therefore the model learns separate {representations} for objects belonging to the same class but appearing on images from different altitudes.
In simple terms: the objects on which the model without Adaptive Resizer did not train can not be recognized during testing.

On the other hand, a  network endowed with the Adaptive Resizer learns representations for every class at 
{one specific scale}. Therefore, the capture-altitude affects detection performance very little as long as every image is resized in the discussed fashion. Essentially, the Adaptive Resizer allows for transferring knowledge in between altitudes. An example: our network can learn from images taken between $0\, m$ and $50\, m$ above ground, and then perform well on images captured between $50\, m$ and $100\, m$.

To prove these claims, we consider four different data set splits for our experiments on height transfer. The construction of these splits is as follows: starting from the above described BEV subsets, we order the images in the data set by their respective capture-altitude.
We then use the 25 \% images with the highest capture-altitude from the training set of the BEV portion as the training set for this task. For the validation set, we use all of the validation images from the BEV subset. Together we call this \textsc{above75}.
Repeating this procedure for the bottom 25 \%, bottom and top 50 \% of the training images yields \textsc{below25}, \textsc{below50}, and \textsc{above50}, respectively. Note that the validation set for each of these splits is the entire validation set of the BEV portion, including all capture-altitudes.

Constructing the data set split this way makes this experiment fit to verify the above claims; if a model performs well on one of the above data set splits, it means that it can generalize from the images it trained on to images with capture-altitudes it never saw before.

Table \ref{tab:resultsdomains} shows that the Adaptive Resizer models consistently outperform their respective baseline counterparts in these experiments. The reported numbers on VisDrone are generally relatively low, as expected, due to the size of the training sets, e.g. \textsc{below25} and \textsc{below75} contain $\approx 300$ training images. Still, in the best case, Adaptive Resizer is three times as good as its baseline ($4.9$ vs $14.2$ AP$^{50}$).

To explain the large improvement in the case of UAVDT, we assume that the baseline's improvement compared to Table \ref{tab:resultsOneStage} comes from UAVDT's gap in between training and validation images we already discussed.
We perceive many more images captured at very high altitudes in its validation set, which do not appear in the training set. The Adaptive Resizer can handle this gap being basically en par with its performance on the whole BEV split of UAVDT, e.g. $47.9$ versus $49.3 \, \text{AP}^{50}$ (see Table \ref{tab:resultsFPN}).

\begin{table}[h]
	\centering
	\begin{tabularx}{\linewidth}{c || c | c || c | c}
		& \multicolumn{2}{c ||}{VisDrone} &  \multicolumn{2}{ c }{UAVDT} \\\hline
		& $D0$ & $D0\, + \,$Adapt. & $D0$ & $D0\, + \,$Adapt. \\\hline\hline
		\textsc{below25} & $5.0$ & $\mathbf{7.2}$ & $9.7$ & $\mathbf{47.9}$ \\\hline
		\textsc{below50} & $7.0$ & $\mathbf{12.0}$ & $26.1$ & $\mathbf{45.5}$ \\\hline
		\textsc{above50} & $4.9$ & $\mathbf{14.2}$ & $32.1$ & $\mathbf{45.4}$ \\\hline
		\textsc{above75} & $8.0$ & $\mathbf{11.2}$ & $18.7$ & $\mathbf{44.5}$
	\end{tabularx}
	\caption{Empirical results for height transfer on VisDrone and UAVDT. Each cell reports the AP$^{50}$ result of either the baseline or adaptive resizer version of an EfficientDet--$D0$.}
	\label{tab:resultsdomains}
	\vspace{-8mm}
\end{table}

\section{Conclusion}
\label{sec:conclusion}

In this work, we proposed a novel preprocessing step. It adjusts the image size according to the height in which the image was captured, solving the scale variance problem in BEV imagery. This method significantly improves detection performance over multiple data sets and object detectors while also improving inference speed, making it applicable to near real-time object detection on mobile platforms. 

We also showed that this method enables object detectors to generalize well to images captured in heights they have never seen before. Furthermore, we used a multi-task fashioned approach to capitalize on our method on generic UAV imagery.

%

{\small
	\bibliographystyle{./IEEEtran}
	\bibliography{refs}
}


\end{document}